\documentclass{article}

\usepackage{arxiv}

\usepackage[utf8]{inputenc} 
\usepackage[T1]{fontenc}    
\usepackage{hyperref}       
\usepackage{url}            
\usepackage{booktabs}       
\usepackage{amsfonts}       
\usepackage{nicefrac}       
\usepackage{microtype}      
\usepackage{lipsum}
\usepackage{graphicx}       %

\title{The Influence of the Other-Race Effect on Susceptibility to Face Morphing Attacks}

\author{
 Snipta Mallick\\
  School of Behavioral and Brain Sciences\\
  The University of Texas at Dallas\\
  Richardson, TX\\
   \And
  Géraldine Jeckeln \\
  School of Behavioral and Brain Sciences\\
  The University of Texas at Dallas\\
  Richardson, TX\\
     \And
Connor J. Parde\\
  School of Behavioral and Brain Sciences\\
  The University of Texas at Dallas\\
  Richardson, TX \\
     \And
Carlos D. Castillo \\
  Whiting School of Engineering \\
  Johns Hopkins University \\
College Park, MD
       \And
 Alice J. O'Toole \\
  School of Behavioral and Brain Sciences\\
  The University of Texas at Dallas\\
  Richardson, TX\\
}

\begin{document}
\maketitle

\begin{abstract}
Facial morphs created between two identities resemble both of the faces used to create the morph. Consequently, humans and machines are prone to mistake morphs made from two identities for either of the  faces used to create the morph. This vulnerability has been exploited in ``morph attacks'' in security scenarios.  Here, we asked whether the ``other-race effect'' (ORE)--- the human advantage for identifying own- vs. other-race faces--- exacerbates morph attack susceptibility for humans. 
We also asked whether face-identification performance in a deep convolutional neural network (DCNN) is
affected by the race of morphed faces. Caucasian (CA) and East-Asian (EA) participants  performed a face-identity matching task on pairs of CA and EA face images in two conditions. In the morph condition, different-identity pairs consisted of an image of identity ``A'' and a 50/50 morph between images of identity ``A'' and ``B''. In the  baseline condition,  morphs of different identities never appeared. 
As expected, morphs were  identified mistakenly more often than original face images. Moreover, CA participants showed an advantage for CA faces in comparison to EA faces (a partial ORE). Of primary interest, morph identification was substantially worse for cross-race faces than for own-race faces. Similar to humans, the DCNN performed more accurately for original face images than for morphed image pairs. Notably, the deep network proved substantially more accurate than humans in both cases. The results point to the possibility that DCNNs might be useful for improving face identification accuracy when morphed faces are presented. They also indicate the significance of the ORE in morph attack susceptibility in applied settings. 
\end{abstract}

\keywords{face morphing \and  face identification\and  face matching \and Deep Convolutional Neural Network \and other-race effect}

\section{Introduction}

Biometrics-based identification and verification systems are deployed widely for a range of security applications, such as border control. Automated Border Control (ABC) e-gates commonly employ face-recognition systems to capture a live image of a traveler to automate passport image authentication using face-image matching \cite{frontex2015best}. If this fails at the live face-recognition stage, the documentation can undergo secondary identity verification by a human border control guard. Accurate identity verification and face matching of travel documentation are critical to determining border-crossing eligibility. This type of face identity matching task with unfamiliar faces is difficult for both human recognizers, such as border control guards, and commercially deployed face-recognition systems \cite{robertson2017fraudulent,robertson2018detecting,kramer2019unfamiliar,ritchie2020face}. The challenges of identity-matching when faces are unfamiliar creates a security vulnerability that can be exploited to bypass ABC e-gates through a face-morphing attack. 

Face morphs have emerged as a new form of identity fraud \cite{ferrara2014magic, Busch_2017_CVPR_Workshops}. In a face-morphing attack, a morphed image, containing a 50/50 average of two identities, is submitted for inclusion in official travel documentation. The live face recognition system can erroneously verify two different individuals for the same passport image. In an applied setting, a criminal actor could morph their face with a similar-looking noncriminal accomplice and subvert ABC e-gates, due to the resemblance of the live face image to the morph. Face-morphing attacks have been determined to be a feasible method of deceiving face-recognition systems at ABC e-gates \cite{ferrara2014magic}. 

Human behavioral studies also indicate that people are susceptible to morph attacks \cite{robertson2017fraudulent,robertson2018detecting, nightingale2021perceptual}. For example, in one study \cite{robertson2017fraudulent}, when participants were given a face-matching task without being warned about computer-generated morphs, 50/50 morphs were accepted as genuine identities at high rates (68 percent). When participants were warned about the presence of morphed images, the false acceptance rate of 50/50 morphs was reduced significantly (21 percent) \cite{robertson2017fraudulent}. 
Moreover, training to identify image artifacts that resulted from morph generation (e.g., overlapping hairlines) improved identity-matching performance \cite{robertson2018detecting}. 
The combined effects of morph detection guidance and training led to higher morph identification rates than just morph detection guidance alone \cite{robertson2018detecting}, although this performance might have been due, in part, to artifact detection \cite{nightingale2021perceptual}. Additionally, individuals who already performed well on distinguishing between two similar-looking faces had better morph identification performance \cite{robertson2018detecting}.

To address the limitations of morph detection, machine-learning based approaches, including some based on Deep-Convolutional Neural Networks (DCNNs), have been leveraged to automate face-morph detection. One early study utilized micro-texture feature extraction and a linear Support Vector Machine (SVM) to determine if a given image was a morph \cite{raghavendra2016detectingmorph}. In comparison to other feature extraction based methods such as Local Binary Patterns-SVM (LBP-SVM) and Local Phase Quantisation-SVM (LPQ-SVM), the SVM used in this study outperformed previous algorithms. Another study combined the features of two DCNNs, VGG-19 and AlexNet, to explore how transfer learning can impact morph detection in digital and print-scanned images \cite{raghavendra2017detecting}. In comparison to the methods used in previous work \cite{raghavendra2016detectingmorph}, such as LBP-SVM and LPQ-SVM, the combined DCNNs performed better on this task. Additionally, a multiple scales attention convolutional neural network (MSA-CNN) trained on morph artifacts outperformed other networks like VGG-19 \cite{parkhi2015deep} and ResNet18 \cite{zhang2022msacnn}. Although these algorithms performed relatively well, it is hard to compare performance due to variability in network designs and morph quality.

As morphing software rapidly improves to produce higher quality images, morph recognition could become even more challenging, due to the reduction of obvious artifacts (e.g. overlapping hairlines). In one recent study \cite{nightingale2021perceptual}, humans and a VGG-based DCNN were tasked with matching identities in pairs of images that included high-quality morphs. These high quality morphs were designed to limit artifacts in the morphing process. Morphs were defined as 50/50 combinations of two identities---one of the identities in the morphed image matched the identity of the other face in the pair. Both humans and machines performed poorly at this task.

Individual variations in susceptibility to morph attacks may be impacted further by the difficulties associated with cross-race face identification (e.g. \cite{chiroro1995investigation, malpass1969recognition, walker2003encoding}). The other-race effect (ORE) describes the findings that humans recognize faces of their ``own'' race more accurately than faces of other races \cite{meissner2001thirty, malpass1969recognition}. Demographic factors such as the race of a face also affect the performance of face-recognition algorithms such as DCNNs \cite{klare2012face, drozdowski2020demographic, cavazos2020accuracy, grother2019face}. Although there are no consistent findings on how race impacts algorithm accuracy, there is clear evidence that algorithm performance can be affected by race-based demographic differences (e.g., \cite{cavazos2020accuracy,grother2019face}). 
In 2019, algorithms submitted to the Face Recognition Vendor Test (FRVT) showed evidence of demographic differences in face-recognition performance \cite{grother2019face}. For example, an algorithm trained on a dataset of immigration application photos had higher false positive rates (erroneous matching of two similar-looking people) for West and East African and East-Asian populations than for Eastern European populations. 

Concerns about algorithm performance across variable demographics are exacerbated in the case of morph attacks, especially in airport or border control settings. The high quality morphs used in \cite{nightingale2021perceptual} included some diverse faces (6 African American/Black, 16 East Asian, 16 South Asian, and 16 Caucasian). The VGG algorithm used in that study was less accurate at identifying morphed images of Black faces than morphed images of East Asian, South Asian, and White faces.  Human identification accuracy as a function of the racial category of the face was not reported.  Although the differences in algorithm performance reported by \cite{nightingale2021perceptual} are of interest, the number/balance of faces across race categories was not controlled enough to provide a direct test of the role of race in the identification of morphed images. 

The goal of this study was to understand how the ORE influences morph attack susceptibility for both humans and a DCNN algorithm. To directly examine this effect, East-Asian (EA) and Caucasian (CA) participants were recruited to complete a face-matching test. The stimuli we used consisted of original images of EA and CA faces and 50 percent morphs of same-race faces (CA-CA and EA-EA morphs). Participants were asked to determine if the faces pictured in image pairs showed the same person or different people. We compared face-matching performance for same-race and other-race morphs and non-morphs (baseline). Participants were unaware that morphed images were present in the test. On the computational side, to compare human and machine performance, a DCNN \cite{ranjan2018crystal} performed the same task as humans on the same stimuli. To minimize the possibility that morphed images could be detected as ``fake'', we presented only the cropped internal regions of the face. The elimination of the external face detail (hair, etc.) also eliminates cues to identity that have been accessible to humans in previous studies. Because most machine-based face-identification algorithms work only on the internal face, this study puts the machine-human comparison on more equal footing than previous comparisons. 


\section{Human face-identification experiment }
\subsection{Methods}
\subsubsection{Design}

The experimental design included three independent variables: participant race (Caucasian, East Asian), face-image race (Caucasian, East Asian), and face-image type (morph, baseline). The latter two varied within-subjects. Accuracy at matching face identity was measured as the area under the receiver operating curve (AUC).

\subsubsection{Participants}
A total of 74 students from the University of Texas at Dallas (UTD) participated in this study. The study was conducted virtually, using Microsoft Teams, due to the social-distancing measures put into practice during the COVID-19 pandemic. Students were recruited using the UTD online sign-up system (SONA) and received one course credit as compensation for their participation. All participants were required to be 18 years of age or older, self-identify as Caucasian or East Asian, and have corrected-to-normal vision.
Race and ethnicity eligibility was determined via self-identification form generated on Qualtrics \cite{qualtrics2019s}. Fourteen participants were excluded due to internet connection instability (data collection impediment). The final data included 60 participants (30 East-Asian participants, 30 Caucasian participants). 
A power analysis using PANGEA \cite{westfall2015pangea} indicated that a total of 60 participants would be sufficient to obtain a power of 0.839 for a medium effect size (\textit{d} = .5). This power analysis was computed to detect a two-way interaction between face-image race (within-subject, East Asian vs Caucasian) and face-image type (within-subject, Baseline vs Morph).\footnote{Note that the design of the power analysis conducted prior to data collection was inaccurate to estimate the sample size required to detect a three-way interaction. A secondary analysis was computed to detect a three-way interaction 
between participant race (between-subject, East Asian vs Caucasian), face-image race (within-subject, East Asian vs Caucasian) and face-image type (within-subject, Baseline vs Morph). Results confirm that a total of 60 participants (30 per participant race group) is sufficient to obtain a power of 0.839 for a medium effect size (\textit{d} = .5).}

\subsubsection{Stimuli}

A total of 64 face-image pairs was used in this experiment. Each face-image pair was assigned to the morph condition (16 East Asian pairs, 16 Caucasian pairs) or the baseline condition (16 East Asian pairs, 16 Caucasian pairs). Both conditions (morph and baseline) included 16 same-identity pairs (two images of the same identity) and 16 different-identity pairs (two images of different identities of the same race, gender, and age group). 

In the morph condition, different-identity pairs included one unedited image (identity A, image 1) and one 50/50 morph between one image of the same identity (identity A, image 2) and one image of a different identity (identity B, image 1). Same-identity pairs were created using one unedited image (identity A, image 1) and one 50/50 morph between two different images of the same identity (identity A, image 2 and image 3). We used morphs in the same-identity pairs to support the Signal Detection Model measures, which require both same- and different-identity pairs in each condition. This ensured also that the performance observed in the morph condition was derived from people's ability to distinguish same- and different-identity pairs, rather than morphed and non-morphed images. All morphed images were cropped around the face to minimize morph artifacts. In the baseline condition, same-identity pairs were created using one unedited image (identity A, image 1) and one cropped image of the same identity (identity A, image 2).  Different-identity pairs included one unedited image (identity A, image 1) and one cropped image of a different person (identity B, image 1). 
See Figure \ref{fig:Experiment} for an example of the stimulus pairs for each condition.

Images were selected from the the Notre Dame Database \cite{schott2010frvt} and showed faces viewed from the front with neutral expressions. The race and gender of the faces in each pair were balanced across the conditions. In each face-image pair, the unedited images consisted of images captured in an uncontrolled illumination setting. All image manipulations (morphing and cropping) were executed on images captured under controlled illumination and performed using the Face Morpher Github repository (Quek, 2019). Additionally, all morphed images underwent further editing with Photoshop and Gimp to remove artifacts (e.g., second irises, smooth appearance, overlapping noses, etc.). Following morphing and cropping, images underwent sharpening in Photoshop to reduce blurred complexions.

\subsubsection{Remote testing protocol}
In order to comply with the COVID-19 social-distancing requirements, human data collection was carried out virtually. The experiment was conducted online using the remote-control features available on Microsoft Teams. All human data were stored locally on the experimenter's computer. The experiment was conducted using PsychoPy v1.84.2 (Peirce, 2007). All participants used Qualtrics survey software to complete the Self-identification survey. 


\subsection{Procedure}
\begin{figure}[h]
  \centering
  \includegraphics[width=6in]{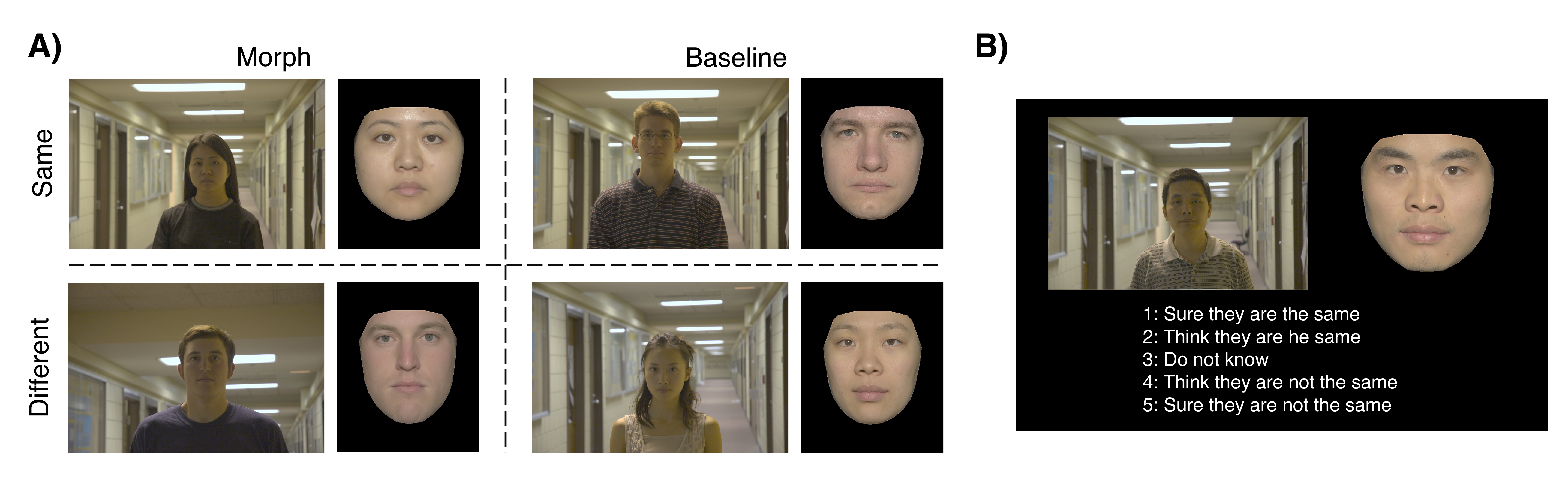}
  \caption{A) Morph condition: Face-image pairs included one unedited image and one cropped 50/50 face morph. The face morphs were created by blending two images of the same identity (\emph{n}= 16) or blending two images of different identities (\emph{n}= 16). Baseline condition: Face-image pairs included one unedited image and one cropped image of the same identity (\emph{n}= 16) or one cropped image of a different identity (\emph{n}= 16). B) Example of a face-matching trial.}
  \label{fig:Experiment}
\end{figure}

\subsubsection{Face-matching task}
All eligible participants received an invitation link to participate in a conference call with the experimenter. The face-matching task was administered virtually using Microsoft Teams. The experiment was conducted locally on the experimenter’s computer. During the experimental session, the subject was given permission to view the experimenters’ screen (via screen sharing) and control the experimenters’ mouse and keyboard remotely. After giving informed control, the participant proceeded with the face-matching task. 

The face-matching test included a total of 64 trials (Fig. \ref{fig:Experiment}B). The face-image pairs in each condition (face-image race, face-image type) were presented in a randomized order. Information pertaining to the experimental conditions (face-image race and face-image type) was not revealed explicitly. On each trial, a face-image pair was presented on the screen. The participants were instructed to determine whether the two images were of the same identity or different identities. Responses were collected using a 5-point certainty scale (1: Sure they are the same; 2: Think they are the same; 3: Do not know; 4: Think they are not the same; 5: Sure they are not the same). The stimuli remained on the screen until a response was entered. 

After completing the face-matching task, the subject was instructed to complete a short demographic survey (see Appendix).

\subsection{Results}

Data were analyzed using a 2 (participant race: East Asian vs Caucasian) x 2 (face-image race: East Asian vs Caucasian) x 2 (face-image type: Baseline vs Morph) mixed-model ANOVA. Face-image race and face-image type were submitted as within-subjects factors and participant race was submitted as a between-subjects factor. The dependent variable (face-matching accuracy) was measured as the AUC. 

\begin{figure}[h]
 \begin{center}
 \includegraphics[width=6in]{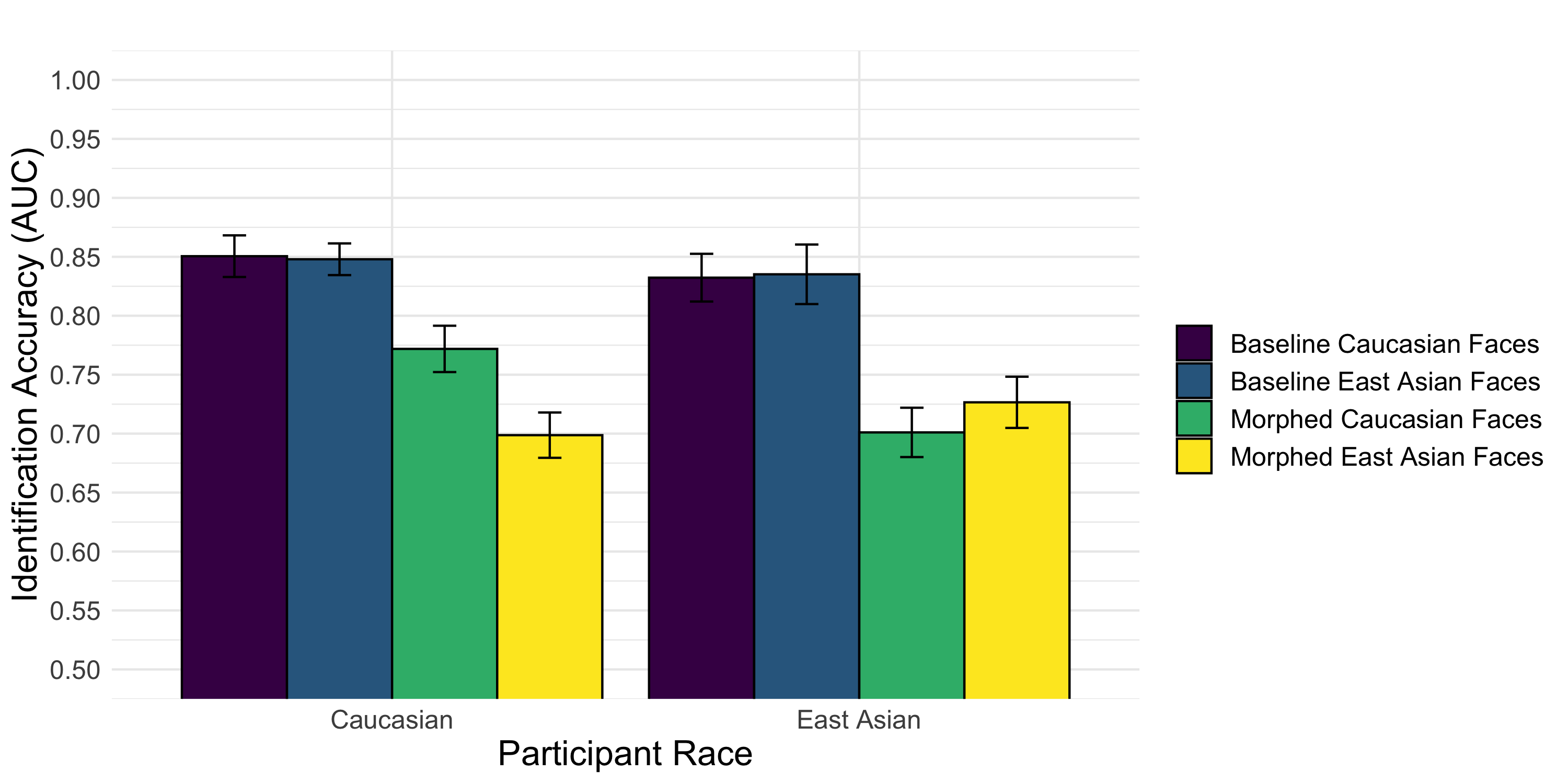}
 \caption{Face-identity matching results. Performance was more accurate for baseline pairs (purple and blue bars) than morph pairs (green and yellow bars). There was a partial ORE with Caucasian Participants performing more accurately on Caucasian pairs than East-Asian pairs, and East-Asian participants performing comparably on both.  Notably, other-race morph pairs 
 proved especially difficult for both East-Asian and Caucasian participants (green and yellow bars).}
 \label{fig:main_results}
 \end{center}
 \end{figure}

As expected, participants performed more accurately 
for the baseline image pairs (\textit{M}= .841, \textit{SE} = .012, 95\% CI [0.818, 0.865]) than the morphed image pairs (\textit{M}= .725,\textit{SE}= .011, 95\% CI [0.703, 0.746]) (see Fig. \ref{fig:main_results}). Specifically, there was a main effect of face-image type (\textit{F}(1,58) = 77.283, \textit{MSe} = 0.011, \textit{p} \textless .001, $\eta_{p}^{2}$ = 0.571). No other main effects were significant.

The results also showed a partial ORE, evidenced by 
a significant interaction between participant race and face-image race (\textit{F}(1,58) = 4.273, \textit{MSe} =0.010 , \textit{p} = 0.043, $\eta_{p}^{2}$ = 0.069). 
A closer look at the data indicates that Caucasian participants were more accurate at identifying Caucasian face pairs (\textit{M}=.811, \textit{SE}=.015, 95\% CI [.780,.842]) than East-Asian face pairs (\textit{M}=.773, \textit{SE}=.016, 95\% CI [.741, .806]). East-Asian participants performed similarly in trials where they viewed East-Asian face pairs (\textit{M}=.781, \textit{SE}=.016, 95\% CI [.748, .813]) and Caucasian face pairs (\textit{M}=.767, \textit{SE}=.015, 95\% CI [.736,.798]). No other two-factor interactions were significant.

Of primary interest for this study, there was  a three-way interaction between participant race, face-image race, and face-image type (\textit{F}(1,58) = 4.49, \textit{MSe}=  0.0073, \textit{p} = 0.038, $\eta_{p}^{2}$ = 0.07). 
Fig. \ref{fig:main_results} shows that 
both the East-Asian and Caucasian participants fared equally well on East-Asian and Caucasian face pairs in the baseline condition. In the morph condition, however, there was an ORE such that East Asians performed more accurately on the East-Asian morph pairs and Caucasians performed more accurately on the Caucasian morph pairs. In other words, the ORE we found was driven strongly by the difficulties participants had with other-race morphs.

In summary, the human experiment replicates the well-documented difficulties people have in matching face identities with morphed stimuli\cite{robertson2017fraudulent,nightingale2021perceptual}. It also showed a partial ORE that was qualified by the three-way interaction. That interaction indicates the additional challenge of face identification with other-race morphs.


\section{DCNN EXPERIMENTS}
\subsection{Methods}
\subsubsection{Network Architecture}
We used a recent high-performing (cf. \cite{Maze2018IARPAJB})  DCNN \cite{ranjan2018crystal} based on the ResNet-101 architecture \cite{he2016deep}. The network contains 101 layers and was trained using the Universe dataset \cite{bansal2017s, ranjan2018crystal}, which contains 5,714,444 images of 58,020 identities. The face-identification accuracy of this network remains high across substantial changes in viewpoint, illumination, and expression. During training, the network used Crystal Loss with the alpha parameter set to 50. Skip connections are used throughout the 101-layered network to retain the amplitude of the error signal. After training was complete, the final fully-connected layer of the network was removed and the output from the penultimate layer (containing 512 units) was used to generate identity descriptors.

We chose this network because it has been used in previous human-machine comparisons with both expert professional forensic face examiners and untrained participants \cite{phillips2018face}. The network performed more accurately than untrained participants and performed at the level of professional forensic face examiners on a challenging face-identification task with a majority of CA faces. The network has been used also to test performance differences between CA and EA faces \cite{cavazos2020accuracy}. In the multi-race tests, overall network performance (AUC) was roughly comparable for EA and CA faces. However, at the low false alarm rates commonly used in security applications, CA faces were identified more accurately than EA faces. 

\subsubsection{Procedure}
The face images used in the human experiment were processed through the DCNN to produce image descriptors.  For each image pair in the human experiment,
the cosine similarity (i.e., normalized dot product) between image descriptors was computed. Higher similarity scores were assumed to indicate higher likelihood that the images showed the same identity. To assess network accuracy, AUC were computed from distributions of similarity scores for the same- and different-identity pairs in each condition.

\subsection{Results}

For image pairs in the baseline (i.e., non-morphed) condition, DCNN face-identification accuracy was perfect (AUC = 1.0). For image pairs in the morph condition, DCNN face-identification accuracy was substantially lower (AUC = 0.891). The decrease in DCNN identification accuracy was more pronounced for Caucasian images (baseline AUC = 1.0; morph AUC = 0.859) than East-Asian images (baseline AUC = 1.0; morph AUC = 0.922).

In summary, the DCNN performed perfectly on the baseline image pairs and less accurately on the morphed image pairs. Furthermore, the DCNN showed an accuracy advantage for EA  over CA morph pairs.

\begin{figure}[h]
 \begin{center}
 \includegraphics[width=6in]{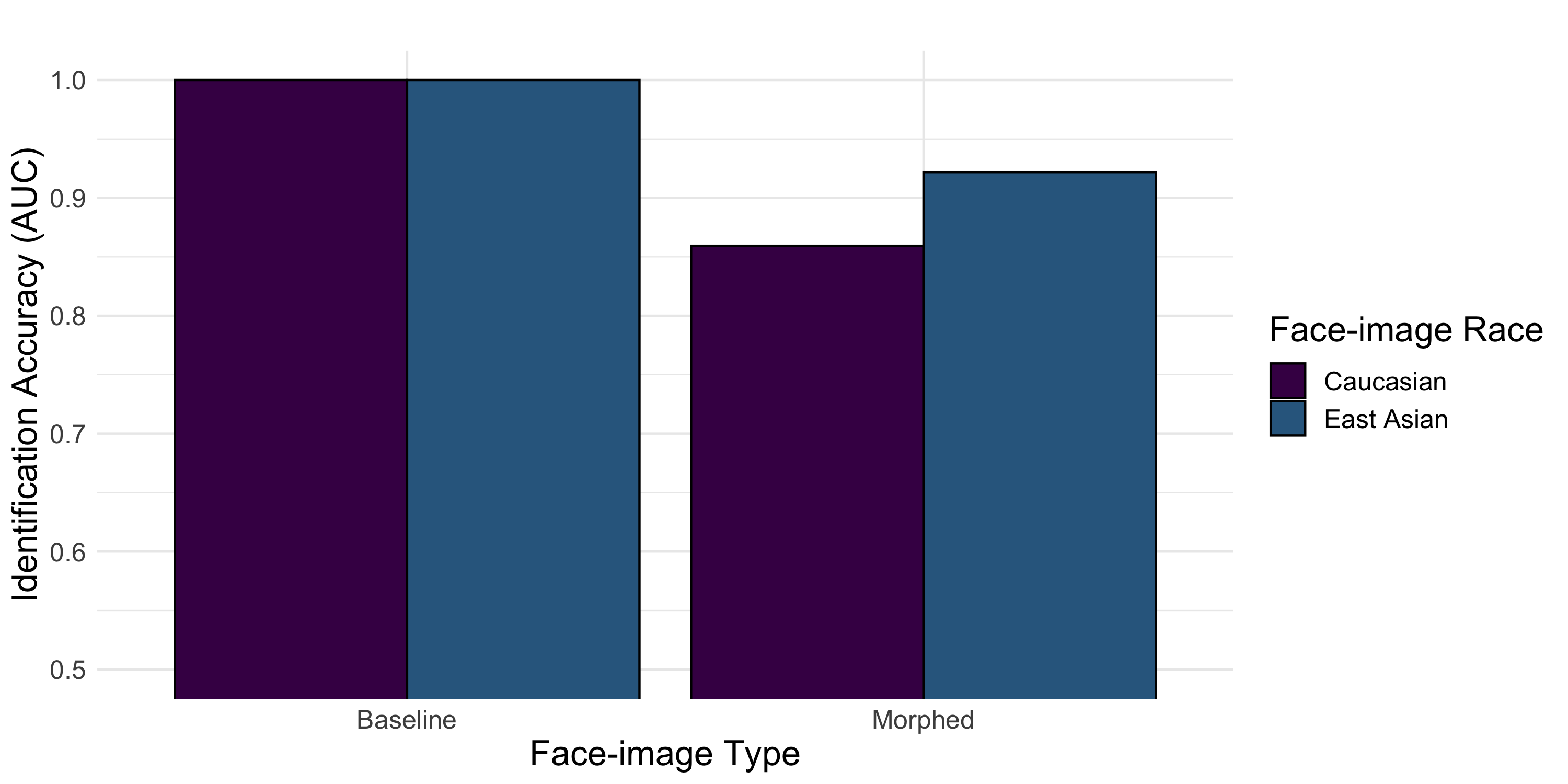}
 \caption{DCNN-based identification accuracy for Caucasian and East-Asian face-image pairs. Accuracy was lower for morphed image pairs in comparison to baseline image pairs, and lower for Caucasian morphed image pairs than East-Asian morphed image pairs.}
 \label{fig:dcnn_accuracy}
 \end{center}
\end{figure}


\section{Summary: Human and Machine Performance}
Both humans and machines showed an advantage for recognizing baseline over morphed images, consistent with previous studies \cite{robertson2017fraudulent}. 
Human participants showed a
partial ORE, driven by East-Asian participants identifying faces in the East-Asian morphs more accurately than in the Caucasian morphs.  Although the DCNN was not tested for a cross-over ORE, the performance of the network was analyzed as a function of the race of the morphed faces. Overall, the performance of the DCNN  surpassed humans on both baseline and morphed image pairs.

\section{Discussion}

The principles underlying the ORE in humans have been well studied \cite{meissner2001thirty, malpass1969recognition}. In response to the emerging security threat posed by face morphs, we analyzed the influence of the ORE on morph attack susceptibility in humans and a DCNN. The present findings expand our understanding of the ORE to include its influence on face identification in a morph-attack scenario. Our human behavioral results demonstrate that the ORE exacerbates the problem of face identification when images are morphed. The DCNN used in this study also performed more accurately than human participants in all cases. Thus, despite its reduced performance for morphed image pairs, and the differences in accuracy for EA and CA faces, the DCNN was always the more accurate face identification ``system''. The findings from this study have significant implications for understanding how the ORE could bias human and algorithmic decision making in border control scenarios. We consider each of these implications in turn.

This is the first study to assess systematically the role of the ORE on morph identification. This was accomplished by conducting a complete cross-over design. Thus, Caucasian and East-Asian participants were tested on CA and EA face-image pairs. 
The present study provides evidence that for morphed images, the ORE can increase morph-attack susceptibility. 
In addition to the use of a cross-over design, the present study provides a more direct test of the ORE on identification with morphs for humans and machines. First, 
we controlled for the possibility that people could detect artifacts in morphed images by cropping the faces to include only the internal face. This also made for a more equitable comparison between the DCNN, which works only on the internal face, and humans. Second, we removed face-image artifacts (e.g., overlapping irises, smooth complexions) that were introduced during the morphing process. Additionally, this study used morphed images pairs for same-identity comparisons to ensure a common ground truth between same- and different-identity comparisons.

The finding that DCNN performance was more accurate for East-Asian than Caucasian morphed-image pairs underscores the unpredictability of algorithms for faces of different races. Although it is clear that the performance of DCNNs is affected by demographics, the source of these effects is less clear and remains an active and open area of research \cite{grother2019face, cavazos2020accuracy}.
Previous studies have indicated that algorithms that originated in China tended to have lower false positive rates on East-Asian faces, although it is not clear whether this difference resulted from training,  optimization, or some other unknown parameter of the algorithms  \cite{grother2019face}. This type of race bias  has been demonstrated also in pre-DCNN algorithms. Earlier algorithms developed in Western countries (e.g. France, Germany, the United States) performed more accurately on Caucasian faces, whereas algorithms developed in East Asian countries (e.g., China, Japan, Korea) performed more accurately on East-Asian faces. Again, however, the source of the effects is not known \cite{phillips2011other}. In the current study, differences in the performance of the  DCNN on EA and CA faces could likewise have resulted from a variety of factors, including imbalances in the training set composition (age, race, etc.), as well as image quality differences \cite{cavazos2020accuracy, grother2019face}. Notwithstanding the demographic effects, the DCNN used in this study fared far better than humans on both baseline and morphed images.





This study lays the groundwork to conduct future assessments for how the ORE could affect morph identification across multiple races. One limitation of this experiment is the consideration of only two racial groups, a limit that can be overcome in future work by expanding the range of racial diversity of participants and face images. 
Concomitantly, there is a wide diversity of demographic effects in DCNNs 
\cite{grother2019face}. Thus, it is incumbent of algorithm users to carefully test and validate the performance of specific algorithms for morphs of different race faces intended for particular applications (e.g., airports in different locations around the world).

The results show that the ORE exacerbates the difficulties associated with morph identification. As fraudsters find new and creative ways of bypassing ABC e-gates, this study elucidates a path forward to mitigate the incidence of morph attacks by investigating how race influences humans and algorithms. The findings have implications for national and international security measures and underscore the complexities of morph detection by humans and algorithms.


\section*{ACKNOWLEDGMENT}
Funding provided by National Eye Institute Grant R01EY029692-03 to AOT and CDC.


\pagebreak
\appendix

\section{Appendix}
\begin{figure}[h]
  \centering
  \includegraphics[width=3in]{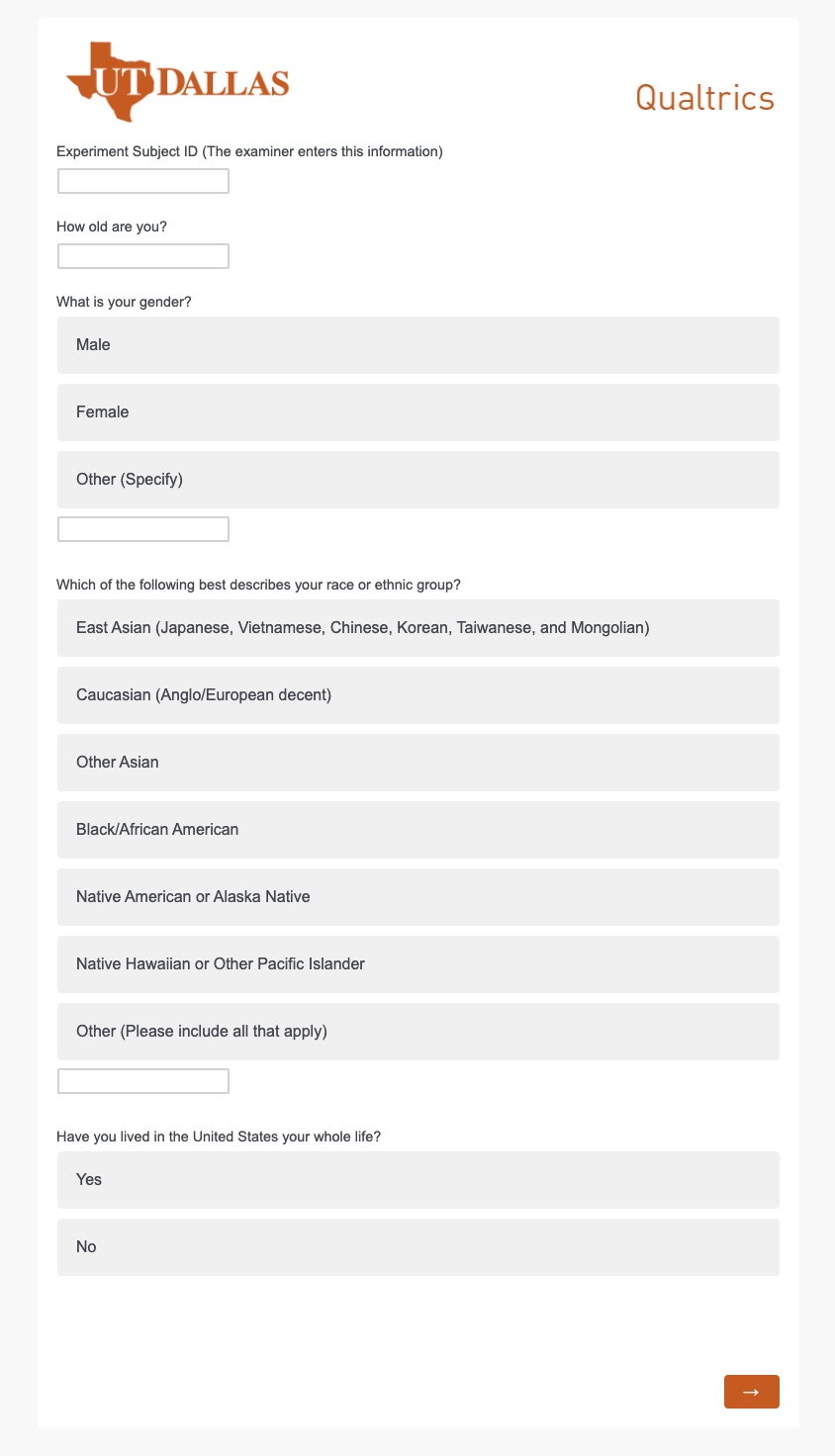}
  \caption{Demographic Survey}
\end{figure}

\label{intro}

\pagebreak

\bibliographystyle{unsrt}  
\bibliography{references}

\end{document}